\documentclass{INTERSPEECH2023}
\usepackage{makecell}
\usepackage{array}

\interspeechcameraready

\title{Text Generation with Speech Synthesis for ASR Data Augmentation}
\name{Zhuangqun Huang*, Gil Keren*, Ziran Jiang*, Shashank Jain*\thanks{*Equal contribution.}, \\ David Goss-Grubbs, Nelson Cheng, Farnaz Abtahi, Duc Le, David Zhang, \\ Antony D'Avirro, Ethan Campbell-Taylor, Jessie Salas, Irina-Elena Veliche, Xi Chen}

\address{Meta AI, USA}
\email{\{teddyhuang, gilkeren, ziranj, shajain, farnazabtahi\}@meta.com}

\begin{document}

\maketitle
 
\begin{abstract}
Aiming at reducing the reliance on expensive human annotations, data synthesis for Automatic Speech Recognition (ASR) has remained an active area of research. While prior work mainly focuses on synthetic speech generation for ASR data augmentation, its combination with text generation methods is considerably less explored. 
In this work, we explore text augmentation for ASR using large-scale pre-trained neural networks, and systematically compare those to traditional text augmentation methods. 
The generated synthetic texts are then converted to synthetic speech using a text-to-speech (TTS) system and added to the ASR training data. 
In experiments conducted on three datasets, we find that neural models achieve 9\%-15\% relative WER improvement and outperform traditional methods. 
We conclude that text augmentation, particularly through modern neural approaches, is a viable tool for improving the accuracy of ASR systems.

\end{abstract}

\section{Introduction}
\label{sec:intro}
Training an end-to-end automatic speech recognition (ASR) model requires thousands of hours of transcribed speech data. However, collecting human annotated speech data is challenging and expensive. Data augmentation using either labelled or unlabelled data \cite{Hori2019, Baskar2021} has been used to alleviate this data scarcity problem. One promising approach is speech data synthesis, which recently contributed to significant progress in domain adaptation \cite{joshi2022simple}, medication names recognition \cite{fazel2021synthasr}, accurate numeric sequences transcription \cite{peyser2019improving}, low-resource languages \cite{zevallos2022data}, 
etc. Most research in this areas has focused on text-to-speech (TTS) generation \cite{hu2022synt++, chen2020improving, rosenberg2019speech, rossenbach2021comparing} and audio augmentation \cite{park2019specaugment, ueno2021data, thai2019synthetic}, and their effects on the resulting ASR accuracy. However, the quality of the texts used for generating synthetic speech data aiming at improving ASR accuracy has not been sufficiently studied. Text augmentation (TA) can potentially greatly affect the downstream ASR accuracy, as one may use it to further diversify and enrich the synthetic speech data. Furthermore, TA methods may also allow quickly setting up an ASR system for a new domain, since a text corpus can be more economically collected, and may be sufficient to reach a target accuracy if effectively augmented. 

The existing literature about TA for improving ASR accuracy is limited. \textit{Rosenberg et. al.} use a non-neural MaxEnt language model (LM) \cite{biadsy2017effectively} for TA and observes a small improvement in word error rate (WER) \cite{rosenberg2019speech}. \textit{Zevallos et. al.} shows a good improvement from the non-augmented dataset \cite{zevallos2022data}. In \cite{zevallos2022data}, synthetic utterances were created by first replacing entities like noun etc. with a delexicalized form which are fed to a seq2seq model to create diverse delexicalized paraphrases. The delexicalized form is replaced by filling in with values found from training data. These methods limit the power of TA for improved ASR and should be replaced with modern, powerful ones pre-trained on large text corpora. In addition, other than LibriSpeech, those works only explore domains where small datasets have simple and short utterances. Furthermore, the reported improvements are convoluted with contributions from other components, like careful TTS optimization with in-domain data.

In this work, we improve over the existing literature by performing TA using a large modern neural network, which benefits from a large-scale corpus for pre-training, for the purpose of ASR data augmentation. In addition, we provide a first of its kind rigorous comparison of modern neural network and rule-based models for the purpose of TA for ASR data augmentation. Specifically, we start from a text corpus, and compare the impact of different TA techniques on different domains with distinct utterance styles. We use both a rule-based non-neural approach (NLX), and two other neural methods. Our neural model based TA methods utilize BART, a large pre-trained model \cite{lewis-etal-2020-bart} which has proven useful in large scale synthetic text generation with minimal human effort \cite{batra2021building}. We utilize mask filling to create synthetic utterances where it can find similar words from the large text corpus it was pre-trained on, unlike \cite{zevallos2022data} which finds words only from the smaller seed data training corpus. These TA methods are compared on the open domain LibriSpeech, a Photo Search domain with short, structured utterances, and a Question and Answering (Q\&A) domain with relatively free-style dictations. To better compare the TA effects, in-domain data from both the Photo Search and Q\&A domains are synthetically generated, and no domain-specific optimization of the TTS systems is conducted. Our key findings and contributions are as follows 
\begin{itemize}
\item TA considerably improves ASR accuracy for open domains or domains with diverse utterances with 10\%-16\% relative WER improvements.
\item Modern neural based TA is superior to the rule-based approach in terms of downstream WER gains for all the three domains investigated. As this finding does not appear in previous literature, it suggests that neural based TA together with a TTS system is an important viable tool for improving ASR accuracy.
\item For domain-specific training, it is possible to start from a small corpus of in-domain text, and modern TA technique should be considered as a standard practice in speech data synthesis for improving ASR systems.
\item Even using only the seed text data for TTS data augmentation with no TA results in significant WER improvements, e.g., relatively $\sim$5\% for open domains and $\sim$15\% for domains with simple, structured utterances.  
\item The model improvement is insensitive to the sizes of augmented data after 8x augmentation, unless reaching a \(>\) 100x augmentation factor. 
\item Combining synthetic data generated from multiple TA methods further improves WER gains for the Q\&A domain, e.g. 11.88\% vs. 15.68\% relatively from the baseline. 
\end{itemize}

\section{Data} \label{sec:data}
\subsection{Datasets} \label{sec:datasets}
We experiment with the LibriSpeech dataset \cite{panayotov2015librispeech} consisting of 960 hours of transcribed training utterances. In this work, both the LibriSpeech test-clean and test-other splits in the evaluation dataset are used for model testing. We also explore two in-house datasets, namely for a Photo Search and a Q\&A ASR domains, where the baseline corpus includes 25k hours of speech from the voice assistant domains and another 125k hours from videos publicly shared by our
Facebook 
users. For the Q\&A domain, we use two internal in-domain test sets that we name "in-domain-1" and "in-domain-2", each containing 6k utterances, originating from live traffic data contributed by internal 
Meta 
employees and from a third-party vendor, respectively. 

The in-domain test set for Photo Search consists of 16.2k utterances from crowdsourcing. For this domain, the instructions in data collection involve issuing utterances to search and retrieve images or query information about them, using either an actual device or a simulator. The collected data is then annotated and marked by humans as success or failure based on the outcome of the query. 
 
\subsection{Text Corpus for Augmentation}
For text generation, we start from a seed corpus that serves as the input to each of the generation methods. For LibriSpeech, the seed corpus contains all the reference text from 281k utterances in the training split. For the Q\&A domain, we use 100k sentences that were gathered by paraphrasing Q\&A user requests from Facebook groups.
The seed corpus for the Photo Search domain contains only 3350 utterances collected using human annotators. The Photo Search domain has well-structured short utterances, e.g. “\textit{show me photos from new year’s}”, while the Q\&A domain contains free-style dictations, e.g., “\textit{could I get some suggestions on ingredients that would go well with white rice to give it a pop of taste}”, etc.

\section{Method} \label{sec:method}
The general pipeline starts from a seed text corpus that is used by a text augmentation method to emit a new corpus of augmented texts. A TTS system then takes the augmented texts and generates synthetic speech utterances. We use a recurrent neural network transducer (RNN-T) model for ASR described in detail in Section \ref{sec:model}. During RNN-T training, both the human annotated data and the augmented TTS data are used, with a hyperparameter controlling their mixing ratio. We train the RNN-T model from scratch for all but the Photo Search domain, where the seed data is small and experiments are done by fine tuning an existing model trained on the data described in Section \ref{sec:datasets}.

\subsection{Text Augmentation}
We compare and contrast both rule-based and model-based approaches to text generation. Early natural language generation (NLG) systems \cite{reiter_dale_2000} utilized content selection, macro / micro planning and surface realization. These were used to create dialog data, whereas our goal is to generate plain text utterances. 
\subsubsection{NLX}
Our rule-based method, called NLX, is a feature-enhanced context-free rule-based system developed in-house. For NLX, a human developer analyzes the seed data to produce syntax rules and uses various tools to mine lexical items such as named entities. The grammar is then used to synthesize a set of utterances. 
\subsubsection{BART}
Our model based approach uses BART \cite{lewis-etal-2020-bart}, a neural network-based model. For BART, we follow the generation scheme from \cite{batra2021building} where the vanilla BART model is first fine-tuned on masked sequences from seed training data which helps with task adaptation similar to the approach in \cite{gururangan-etal-2020-dont}. During generation we run a mask filling approach to generate synthetic texts. We use two kinds of masking strategies: 1) \textit{Random Masking} (BART R.M.), where we insert masks to seed data randomly. 2) \textit{Custom Masking} (BART C.M.), where we use Part of Speech tags to identify candidate words for masking like nouns and verbs, and replace them one at a time with masks. 

\begin{table*}[]
\small
\caption{Samples of synthetic texts generated from a fine-tuned BART model.}
\label{table:table3}
\centering
\begin{tabular}{|l|l|l|}
\hline
                                 & \thead{\textbf{Random Masking}} & \thead{\textbf{Custom Masking}}  \\ [2pt]
\hline

\textbf{Seed Text} & does anyone have a travel discount code please?  & does anyone have a \textbf{\(<\)mask\(>\)} code please?  \\ [2pt] 

\hline
\textbf{Augmented Texts} & \makecell[l]{ \\[-1em]
is anyone selling a travel discount logo please? \\ [2pt]
do you have a travel discount app please? \\[2pt]
could anyone post a restaurant discount code please? \\ [2pt] 
\\[-1em]}  & \makecell[l]{
\\[-1em]
does anyone have a discount code please? \\ [2pt]
does anyone have a delivery code please? \\ [2pt]
does anyone have a spare code please? \\ [2pt]
\\[-1em]
} \\
\hline

\end{tabular}
\end{table*}

\subsubsection{Quality and Throughput}
For NLX, text quality depends on the developed grammar rules. A developer with linguistic expertise can develop a precise grammar for the task. For highly constrained domains, a grammar with acceptable coverage can be developed in a matter of a few days. This is more challenging for complex and open domains, such as LibriSpeech, and requires more efforts. Furthermore, while some utterances tend to be syntactically correct, they may be less semantically common. Below are examples of synthetic texts from NLX:
\begin{itemize}
\item LibriSpeech: \textit{those hands could have been mexican}
\item Photo Search: \textit{show me all pics of these this Tuesday}
\item Q\&A: \textit{search in burpee gardening for fences}
\end{itemize}

For BART, as shown by \textit{Batra et. al.} \cite{batra2021building}, random masking results in highly diverse utterances. For custom masking, we get less variations as we mask only nouns, verbs one at a time as compared to inserting random masks repeatedly N times. Table \ref{table:table3} shows some samples generated from a fine-tuned BART model. The BART method is free from the linguistic expertise and human intervention that NLX does. It generates millions of high-quality utterances within 2 days using a single GPU.

\subsection{TTS} \label{sec:TTS}
The TTS pipeline consists of a linguistic frontend and an acoustic backend. The frontend takes plain text as input and emits its phonetic representation (grapheme to phoneme) and an additional prosodic information for each token in the output sequence. These are fed into the acoustic backend that consists of a transformer-based prosody model which predicts phone-level F0 values and duration, and a transformer-based spectral model that produces frame-level mel-cepstral coefficients, F0 and periodicity features, and finally a sparse WaveRNN-based neural vocoder that predicts the final waveform \cite{wu2021transformer}.

The acoustic model is trained on a multi-speaker dataset consisting of 170 hours of TTS-quality audio across 96 English speakers, while the WaveRNN vocoder was trained for each speaker separately. We augment the 96 voices with 3 speeds and 5 pitch variations. This TTS system is fixed across all experiments.
 
\subsection{Audio Augmentation}
The audio augmentation pipeline includes speed distortion to introduce speed diversity by factors of 0.9, 1.0 and 1.1 for 40\%, 20\% and 40\% of the data, respectively. In addition, a pool of 1.72M noise tracks are used. We keep 40\% of audios noise-free, nearly 60\% with one noise added, and \(<\) 1\% having 2-4 noises. The SNR has an \textit{N}(12.50, 17.31) distribution. 
 
\subsection{ASR RNN-T Model} \label{sec:model}
Our ASR model is a low-latency streamable RNN-T model, following the recipe in \cite{Le2021}. 
Its audio encoder consists of 20 Emformer layers \cite{shi2021emformer}. The RNN predictor has 3 LSTM layers followed by a linear layer, and layer normalization in between every two other layers. The joiner module is simply a ReLU activation function followed by a single linear layer. We tokenize all texts to 5000 subword units \cite{DBLP:conf/emnlp/KudoR18} for LibriSpeech, and 4096 units for the two in-house datasets. In total, our model has 78M trainable parameters for LibriSpeech, and 76M for the in-house datasets. Adam optimizer is used for updating the model parameters during training.

\section{Experiments}
We train ASR models for the three datasets described above. 
For each dataset, we compare the baseline model trained with no data augmentation to models with augmented TTS utterances. 
The TTS utterances are either created from the seed text utterances, the rule-based NLX method, the neural BART method with either random or custom masking, or a combination of the above. For all data augmentation methods, the seed text is always included in the generated TTS utterances. 

\subsection{Generated Data}
The seed text data for the LibriSpeech, Photo Search and Q\&A datasets is comprised of 281k, 3350, and 100k utterances, respectively. Both NLX and BART methods use an augmentation factors of 8.2x, 300x, and 8.0x, which resulted in 2.3M, 1.0M and 800k utterances respectively for the three datasets above. Those were converted to speech utterances using the TTS system described in Section \ref{sec:TTS}.
In order to simulate a situation of no human annotated utterances from a target speech domain, all Photo Search and Q\&A in-domain training data is synthetically generated from our TTS system. When adding the synthetic texts to the training corpus, we control the mixing ratio, i.e., how often this data is seen by the model compared to other training data. We tune this hyperparameter for each data and experiment separately. We found that higher mixing ratios are acceptable for LibriSpeech, such that up to 50\% of the data seen is synthetic. For in-house datasets, we found that lower mixing ratios, up to 10\%, are optimal for best WER. 

\begin{table*}[]
\small
\caption{Word error rate results (in percents) for baseline models and different data augmentation methods.}
\label{table:table1}
\centering
\begin{tabular}{|l|cc|c|cc|}
\hline
\multicolumn{1}{|l|}{}            & \multicolumn{2}{c|}{\textbf{LibriSpeech}}                          & \textbf{Photo Search} & \multicolumn{2}{c|}{\textbf{Q\&A}}                                   \\ \hline
\textbf{Seed Data}                & \multicolumn{2}{c|}{281 k}                                         & 3350                  & \multicolumn{2}{c|}{100 k}                                           \\ \hline
\makecell[l]{\textbf{Augmented Data}}           & \multicolumn{2}{c|}{2.3 M}                                         & 1.0 M                 & \multicolumn{2}{c|}{800 k}                                           \\ \hline
\textbf{Test Set}                 & \multicolumn{1}{c|}{test-other}   & test-clean   & in-domain    & \multicolumn{1}{c|}{in-domain-1} & in-domain-2 \\ \hline
\textbf{Baseline}             & \multicolumn{1}{c|}{9.02}                  & 3.52                  & 4.61                  & \multicolumn{1}{c|}{9.13}                   & 7.91                   \\ \hline
\makecell[l]{\textbf{Seed Texts}}           & \multicolumn{1}{c|}{8.53}                  & 3.34                  & \textbf{3.94}                  & \multicolumn{1}{c|}{8.96}                   & 7.78                   \\ \hline
\textbf{NLX}                  & \multicolumn{1}{c|}{8.65}                  & 3.31                  & 4.18                  & \multicolumn{1}{c|}{8.34}                   & 7.15                   \\ \hline
\textbf{BART C.M.} & \multicolumn{1}{c|}{8.51}                  & 3.24                  & \textbf{3.94}                  & \multicolumn{1}{c|}{8.41}                   & 7.15                   \\ \hline
\textbf{BART R.M.} & \multicolumn{1}{c|}{\textbf{8.14}}                  & \textbf{3.13}                  & \textbf{3.94}                  & \multicolumn{1}{c|}{8.32}                   & 7.07                   \\ \hline

\makecell[l]{\textbf{NLX + BART R.M.}} & \multicolumn{1}{c|}{8.36}                  & 3.14                  & 4.03                  & \multicolumn{1}{c|}{8.08}                   & 6.78                   \\ \hline

\makecell[l]{\textbf{NLX + BART R.M. + BART C.M.}} & \multicolumn{1}{c|}{8.26}                  & 3.18                  & 3.99                  & \multicolumn{1}{c|}{\textbf{7.97}}                   & \textbf{6.67}                   \\ \hline
\end{tabular}

\end{table*}

\subsection{Results and Discussion}
Table \ref{table:table1} lists the WERs for different trained models.
\subsubsection{Seed Data without Augmentation}
By introducing TTS from the seed texts only, the WERs for all the three domains are reduced in different magnitudes, where Q\&A sees a trivial relative improvement (\(<\) 2\%), LibriSpeech achieves 5.2\%-5.4\%, and Photo Search achieves 14.53\%. Note that we only used 3350 utterances for the Photo Search domain. Thus, for domain with simple and structured utterances, a few thousands of text utterances are sufficient to achieve high performance ASR with a WER \(<\) 4\%. TTS from seed texts is expected to improve both photo search and Q\&A, as the baseline model was trained without in-domain data. Interestingly, we only observed the improvement on the simple photo search domain. 

\subsubsection{Augmentation with BART or NLX}
Adding synthetic texts to the seed data managed to improve WER results in many but not all settings. Below we compare models trained with TA to models trained with the seed texts only. NLX leads to a regression for both LibriSpeech and the Photo Search domain, while a significant relative improvement (6.1\%-8.1\%) for the Q\&A domain is observed. For BART C.M., the WER for the Photo Search domain has no regression, and sees improvements for the other two domains. These improvements are further increased with BART R.M. 

A few interesting insights arise from those comparisons. First, We see that overall the BART neural approaches result in better WER gains compared to the rule-based NLX. While this result may not be surprising, the superiority of neural methods in TA followed by TTS for ASR data augmentation was not yet verified in the existing literature. Second, we see that TA improves WER only for the two datasets that have a large and diverse seed corpus, suggesting that TA has a limited power when the seed data is small and structured. 

\subsubsection{Combining NLX and BART}
We also experiment with the combination of synthetic data from different TA approaches, where data is simply mixed together. We find that combining data sources that result in gains individually (over just using the seed texts) results in larger gains. For example, for the Q\&A domain, all TA methods improved over the seed data, and the ASR accuracy constantly improves from BART R.M. to BART R.M. + NLX, to BART R.M. + NLX + BART C.M. However, for LibriSpeech and the Photo Search domain, NLX does not result in large gains, and we do not observe improvement after combining it with other data sources. 

\subsubsection{Effects of Augmented Factors}
To further examine the effect of the size of augmented text corpus, Table \ref{table:table2} lists additional WERs from both test-clean and test-other where different sizes of BART R.M. texts are introduced. From 2.3M to 7.3M, equivalent to an augmentation factor of 8x to 26x, the WERs are fairly stable with a fluctuation in relative improvement of less than 1\%. We also augmented our data by 103x and achieved WER improvements, where the test-clean set WER drops from 3.11\% to 3.00\%. Similarly, for Q\&A we also augmented the texts to 8.3M by 83x augmentation with BART R.M., but obtained no additional gains. Those results indicate that in some cases TA methods may saturate and show no further gains after a certain number of utterances are generated. 

Note that synthetic data is generated at scale and at speed in this work. Thus, we are less focused on small augmentation factors. Instead, for both Q\&A and photo search, we choose factors to introduce ~4k hrs of audio or 2.7\% of the training data by duration. We also control TTS data used during training by the mixing ratio. Both are to avoid overfitting to specific domains. For LibriSpeech, we use a factor similar to the Q\&A domain. The effect of TTS data size on different domains requires further investigation in future work.

\subsubsection{Other Factors}
The size of the seed data plays a critical role, e.g. 10k (instead of 100k) Q\&A seed texts only achieves \(<\) 5\% relative improvement. Efforts to collect high-quality seed text corpus are non-trivial, though dramatically reduced comparing with collection of annotated paired audio-text data. Given the large amounts of raw data available from external sources, e.g. web data, one might argue that TA might not be needed. However, this type of data always requires processing for privacy compliance, noise acceptance, and right distributions for target domains. In this regard, it is highly desired to both increase lexical diversities while reducing seed corpus size and to explore the effective use of data from other resources. 

In this work, audio augmentation with increased numbers of TTS voices and audio distortions \cite{fazel2021synthasr}, e.g. noise addition and speed and pitch augmentation, are beneficial for ASR improvement. Instead of 96 voices with audio distortion used in this work, if we use only 12 voices and no audio distortion, the best relative improvement that can be achieved is \(<\) 4\%. In future work, we will be increasing voice diversity by enlarging our pool of TTS voices, and enriching our customizable TTS with more personalized elements. 

During the RNN-T decoding, external LMs are not used in this work, but only the standard RNN-T beam search is employed. Nonetheless, comparing the effectiveness of TTS vs. external LMs in utilizing the synthesized text is a natural area for further exploration.


\begin{table}[]
\small
\caption{LibriSpeech, Word Error Rate [\%] vs. Text Size.}
\label{table:table2}
\resizebox{\columnwidth}{!}{\begin{tabular}{|l|c|c|c|c|}
\hline
                                 & \textbf{4.5 M} & \textbf{5.9 M} & \textbf{7.3 M} & \textbf{29 M} \\ \hline
\textbf{Augmentation Factor} & 16x            & 21x          & 26x           & 103x             \\ \hline
\textbf{test-other} & 8.21           & 8.18           & 8.21           & \textbf{8.10}           \\ \hline
\textbf{test-clean} & 3.10            & 3.11           & 3.11           & \textbf{3.00}             \\ \hline
\end{tabular}}
\end{table}

\section{Conclusion}
In this work, we focus on text augmentation for speech data synthesis, which we show to be an effective approach for improving ASR systems. 
Compared with the rule-based method, a neural approach based on modern powerful pre-trained language models is highly effective for synthetic text generation and the subsequent ASR model training.  
Our results show that for the open-domain LibriSpeech and Q\&A datasets, neural text augmentation based on a fine tuned BART model improves the relative performance by 9.76\% - 15.68\% compared to the baselines.
However, text augmentation results in no gain for simple domains with highly-structured utterances.  
Furthermore, the additional WER gains by simply mixing together data from different TA approaches confirmed synergy among data from these techniques. 
This work suggests that it could be cost-effective to improve an ASR system using an in-domain text corpus instead of an expensive human-annotated audio corpus, and modern text generation techniques should be considered as an important element in speech data synthesis for improving ASR systems. 

\section{Acknowledgements}
We would like to thank Xiao Yang, Jiale Zhi, Thilo Koehler, Didi Zhang, Azadeh Nematzadeh, Qing He, Jilong Wu, and Tammy Stark for their help in data collection, TTS, data augmentation, and helpful discussions and suggestions.

\bibliographystyle{IEEEtran}
\bibliography{mybib}

\begin{thebibliography}{10}
\providecommand{\url}[1]{#1}
\csname url@samestyle\endcsname
\providecommand{\newblock}{\relax}
\providecommand{\bibinfo}[2]{#2}
\providecommand{\BIBentrySTDinterwordspacing}{\spaceskip=0pt\relax}
\providecommand{\BIBentryALTinterwordstretchfactor}{4}
\providecommand{\BIBentryALTinterwordspacing}{\spaceskip=\fontdimen2\font plus
\BIBentryALTinterwordstretchfactor\fontdimen3\font minus
  \fontdimen4\font\relax}
\providecommand{\BIBforeignlanguage}[2]{{%
\expandafter\ifx\csname l@#1\endcsname\relax
\typeout{** WARNING: IEEEtran.bst: No hyphenation pattern has been}%
\typeout{** loaded for the language `#1'. Using the pattern for}%
\typeout{** the default language instead.}%
\else
\language=\csname l@#1\endcsname
\fi
#2}}
\providecommand{\BIBdecl}{\relax}
\BIBdecl

\bibitem{Hori2019}
T.~Hori, R.~Astudillo, T.~Hayashi, Y.~Zhang, S.~Watanabe, and J.~L. Roux,
  ``Cycle-consistency training for end-to-end speech recognition,'' in
  \emph{ICASSP}, 2019.

\bibitem{Baskar2021}
M.~K. Baskar, L.~Burget, S.~Watanabe, R.~F. Astudillo, and J.~H. Černocký,
  ``Eat: Enhanced asr-tts for self-supervised speech recognition,'' in
  \emph{ICASSP}, 2021.

\bibitem{joshi2022simple}
R.~Joshi and A.~Singh, ``A simple baseline for domain adaptation in end to end
  asr systems using synthetic data,'' in \emph{ECNLP}, 2022, pp. 244--249.

\bibitem{fazel2021synthasr}
A.~Fazel, W.~Yang, Y.~Liu, R.~Barra{-}Chicote, Y.~Meng, R.~Maas, and J.~Droppo,
  ``Synthasr: Unlocking synthetic data for speech recognition,'' in
  \emph{Interspeech}, 2021, pp. 896--900.

\bibitem{peyser2019improving}
C.~Peyser, H.~Zhang, T.~N. Sainath, and Z.~Wu, ``Improving performance of
  end-to-end {ASR} on numeric sequences,'' in \emph{Interspeech}, G.~Kubin and
  Z.~Kacic, Eds., 2019, pp. 2185--2189.

\bibitem{zevallos2022data}
R.~Zevallos, N.~Bel, G.~C{\'a}mbara, M.~Farr{\'u}s, and J.~Luque, ``Data
  augmentation for low-resource quechua asr improvement,'' \emph{arXiv preprint
  arXiv:2207.06872}, 2022.

\bibitem{hu2022synt++}
T.-Y. Hu, M.~Armandpour, A.~Shrivastava, J.-H.~R. Chang, H.~Koppula, and
  O.~Tuzel, ``Synt++: Utilizing imperfect synthetic data to improve speech
  recognition,'' in \emph{ICASSP}, 2022, pp. 7682--7686.

\bibitem{chen2020improving}
Z.~Chen, A.~Rosenberg, Y.~Zhang, G.~Wang, B.~Ramabhadran, and P.~J. Moreno,
  ``Improving speech recognition using gan-based speech synthesis and
  contrastive unspoken text selection.'' in \emph{INTERSPEECH}, 2020, pp.
  556--560.

\bibitem{rosenberg2019speech}
A.~Rosenberg, Y.~Zhang, B.~Ramabhadran, Y.~Jia, P.~Moreno, Y.~Wu, and Z.~Wu,
  ``Speech recognition with augmented synthesized speech,'' in \emph{2019 IEEE
  automatic speech recognition and understanding workshop (ASRU)}, 2019, pp.
  996--1002.

\bibitem{rossenbach2021comparing}
N.~Rossenbach, M.~Zeineldeen, B.~Hilmes, R.~Schl{\"u}ter, and H.~Ney,
  ``Comparing the benefit of synthetic training data for various automatic
  speech recognition architectures,'' in \emph{2021 IEEE Automatic Speech
  Recognition and Understanding Workshop (ASRU)}, 2021, pp. 788--795.

\bibitem{park2019specaugment}
D.~S. Park, W.~Chan, Y.~Zhang, C.~Chiu, B.~Zoph, E.~D. Cubuk, and Q.~V. Le,
  ``Specaugment: {A} simple data augmentation method for automatic speech
  recognition,'' in \emph{Interspeech}, G.~Kubin and Z.~Kacic, Eds., 2019, pp.
  2613--2617.

\bibitem{ueno2021data}
S.~Ueno, M.~Mimura, S.~Sakai, and T.~Kawahara, ``Data augmentation for asr
  using tts via a discrete representation,'' in \emph{2021 IEEE Automatic
  Speech Recognition and Understanding Workshop (ASRU)}, 2021, pp. 68--75.

\bibitem{thai2019synthetic}
B.~Thai, R.~Jimerson, D.~Arcoraci, E.~Prud'hommeaux, and R.~Ptucha, ``Synthetic
  data augmentation for improving low-resource asr,'' in \emph{2019 IEEE
  Western New York Image and Signal Processing Workshop (WNYISPW)}, 2019, pp.
  1--9.

\bibitem{biadsy2017effectively}
F.~Biadsy, M.~Ghodsi, and D.~Caseiro, ``Effectively building tera scale maxent
  language models incorporating non-linguistic signals,'' in
  \emph{Interspeech}, 2017, pp. 2710--2714.

\bibitem{lewis-etal-2020-bart}
M.~Lewis, Y.~Liu, N.~Goyal, M.~Ghazvininejad, A.~Mohamed, O.~Levy, V.~Stoyanov,
  and L.~Zettlemoyer, ``{BART}: Denoising sequence-to-sequence pre-training for
  natural language generation, translation, and comprehension,'' in \emph{ACL},
  2020, pp. 7871--7880.

\bibitem{batra2021building}
S.~Batra, S.~Jain, P.~Heidari, A.~Arun, C.~Youngs, X.~Li, P.~Donmez, S.~Mei,
  S.~Kuo, V.~Bhardwaj \emph{et~al.}, ``Building adaptive acceptability
  classifiers for neural nlg,'' in \emph{EMNLP}, 2021, pp. 682--697.

\bibitem{panayotov2015librispeech}
V.~Panayotov, G.~Chen, D.~Povey, and S.~Khudanpur, ``Librispeech: an asr corpus
  based on public domain audio books,'' in \emph{ICASSP}, 2015, pp. 5206--5210.

\bibitem{reiter_dale_2000}
E.~Reiter and R.~Dale, \emph{Building Natural Language Generation Systems},
  ser. Studies in Natural Language Processing.\hskip 1em plus 0.5em minus
  0.4em\relax Cambridge University Press, 2000.

\bibitem{gururangan-etal-2020-dont}
S.~Gururangan, A.~Marasovi{\'c}, S.~Swayamdipta, K.~Lo, I.~Beltagy, D.~Downey,
  and N.~A. Smith, ``Don{'}t stop pretraining: Adapt language models to domains
  and tasks,'' in \emph{ACL}, 2020, pp. 8342--8360.

\bibitem{wu2021transformer}
C.~Wu, Z.~Xiu, Y.~Shi, O.~Kalinli, C.~Fuegen, T.~Koehler, and Q.~He,
  ``Transformer-based acoustic modeling for streaming speech synthesis.'' in
  \emph{Interspeech}, 2021, pp. 146--150.

\bibitem{Le2021}
D.~Le, M.~Jain, G.~Keren, S.~Kim, Y.~Shi, J.~Mahadeokar, J.~Chan, Y.~Shangguan,
  C.~Fuegen, O.~Kalinli, Y.~Saraf, and M.~L. Seltzer, ``Contextualized
  streaming end-to-end speech recognition with trie-based deep biasing and
  shallow fusion,'' in \emph{Interspeech}, 2021, pp. 1772--1776.

\bibitem{shi2021emformer}
Y.~Shi, Y.~Wang, C.~Wu, C.-F. Yeh, J.~Chan, F.~Zhang, D.~Le, and M.~Seltzer,
  ``Emformer: Efficient memory transformer based acoustic model for low latency
  streaming speech recognition,'' in \emph{ICASSP}, 2021, pp. 6783--6787.

\bibitem{DBLP:conf/emnlp/KudoR18}
T.~Kudo and J.~Richardson, ``Sentencepiece: {A} simple and language independent
  subword tokenizer and detokenizer for neural text processing,'' in
  \emph{EMNLP}, E.~Blanco and W.~Lu, Eds., 2018, pp. 66--71.

\end{thebibliography}

\end{document}